\documentclass[10pt,twocolumn,letterpaper]{article}

\usepackage{cvpr}              

\usepackage[dvipsnames]{xcolor}

\usepackage[accsupp]{axessibility} 

\usepackage{graphicx}
\usepackage{amsmath}
\usepackage{amssymb}
\usepackage{booktabs}
\usepackage{bbm}
\usepackage{multirow}
\usepackage{pifont}
\usepackage{amssymb}
\usepackage{overpic}
\usepackage{makecell}

\DeclareMathOperator*{\argmin}{arg\,min}

\definecolor{cvprblue}{rgb}{0.21,0.49,0.74}
\usepackage[pagebackref,breaklinks,colorlinks,citecolor=cvprblue]{hyperref}

\usepackage[capitalize]{cleveref}
\crefname{section}{Sec.}{Secs.}
\Crefname{section}{Section}{Sections}
\Crefname{table}{Table}{Tables}
\crefname{table}{Tab.}{Tabs.}
 
\newcommand{\cmark}{\ding{51}}%
\newcommand{\xmark}{\ding{55}}%

\def\paperID{4898} 
\def\confName{CVPR}
\def\confYear{2024}

\title{Fully Geometric Panoramic Localization}

\author{Junho Kim\textsuperscript{1}, Jiwon Jeong\textsuperscript{3}, and Young Min Kim\textsuperscript{1, 2}
\and {\small \phantom{ }} \vspace{-1em}\\
\textsuperscript{1} {\small Dept. of Electrical and Computer Engineering, Seoul National University} \\
\textsuperscript{2} {\small Interdisciplinary Program in Artificial Intelligence and INMC, Seoul National University} \\
\textsuperscript{3} {\small Dept. of Electrical Engineering, Stanford University}
}

\begin{document}
\maketitle
\begin{abstract}
We introduce a lightweight and accurate localization method that only utilizes the geometry of 2D-3D lines.
Given a pre-captured 3D map, our approach localizes a panorama image, taking advantage of the holistic $360^\circ$ view.
The system mitigates potential privacy breaches or domain discrepancies by avoiding trained or hand-crafted visual descriptors.
However, as lines alone can be ambiguous, we express distinctive yet compact spatial contexts from relationships between lines, namely the dominant directions of parallel lines and the intersection between non-parallel lines.
The resulting representations are efficient in processing time and memory compared to conventional visual descriptor-based methods.
Given the groups of dominant line directions and their intersections, we accelerate the search process to test thousands of pose candidates in less than a millisecond without sacrificing accuracy.
We empirically show that the proposed 2D-3D matching can localize panoramas for challenging scenes with similar structures, dramatic domain shifts or illumination changes.
Our fully geometric approach does not involve extensive parameter tuning or neural network training, making it a practical algorithm that can be readily deployed in the real world.
Project page including the code is available through this link: \url{https://82magnolia.github.io/fgpl/}.

\end{abstract}    
\section{Introduction}
\label{sec:intro}
\begin{figure}[t]
  \centering
    \includegraphics[width=\linewidth]{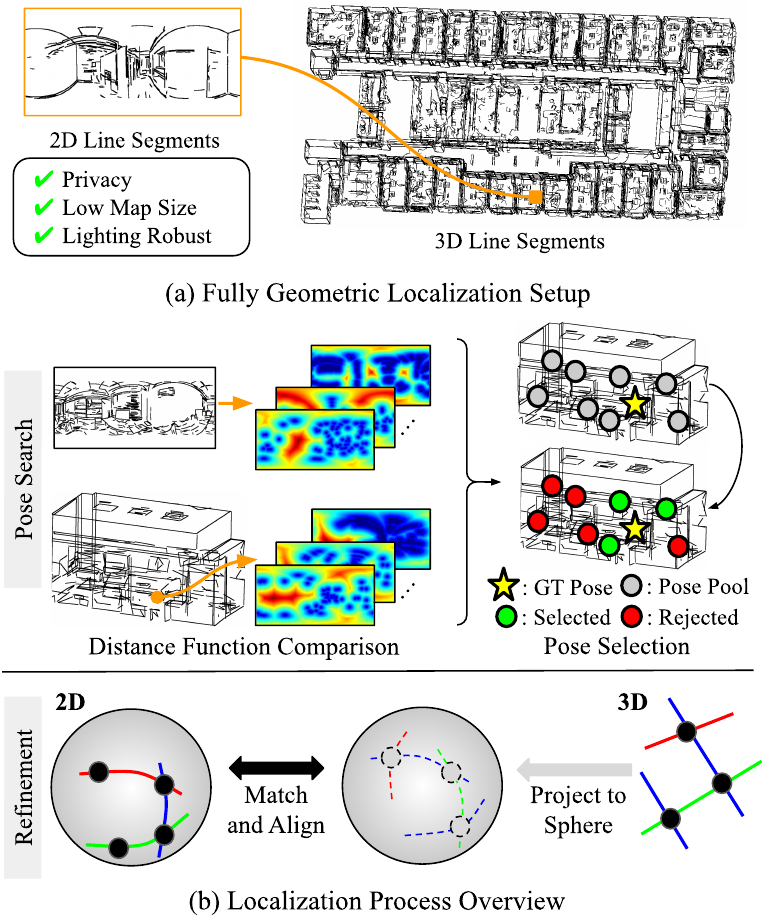}
   \caption{Overview of our method. (a) We target fully geometric localization using panoramas, where we only exploit lines in 2D and 3D. (b) Our method first searches for promising poses by comparing point and line distance functions that describe the holistic distribution of lines and their intersections. Then we refine each selected pose by aligning the line intersections on the sphere.}
   \label{fig:overview}
\vspace{-2em}
\end{figure}

Visual localization considers the problem of estimating the camera pose with respect to a 3D map using an input image.
When designing a visual localization system, the key desiderata is defining the \textit{3D map} and the associated \textit{image features} to match against the map.
The most common choice in modern localization systems is to use 3D maps from Structure-from-Motion (SfM)~\cite{colmap_1, colmap_2} or dense laser scans~\cite{inloc} and match an image against extracted visual descriptors~\cite{netvlad,superpoint,sift}.
While being performant, such a design choice has several limitations~\cite{gomatch}: (i) building and storing the map of image features can be costly, (ii) privacy breaches may occur in client-server localization scenarios as visual descriptors cannot securely hide clients' view~\cite{privacy_affine,pittaluga_revealing,privacy_line2d}, and (iii) localization may fail in challenging conditions such as lighting or scene changes.

In this paper, we argue that the holistic context of lines, when properly exploited, can be sufficient for performing accurate and scalable panoramic localization.
To elaborate, as shown in Figure~\ref{fig:overview}, we propose to \textit{solely} use 3D lines as the map, and 2D lines extracted from the query panorama as features for localization.
Thanks to the large field of view, localization methods using panoramas~\cite{sphere_cnn,gosma,piccolo,cpo,ldl} are more robust against scene changes or repetitive structures compared to regular cameras.
Further, the line maps can resolve the drawbacks of conventional descriptors:
they are much cheaper to store than SfM or dense point maps, preserve user privacy as lines alone lack learnt/photometric features, and are robust to illumination changes.

From lines, we devise \textit{fully geometric} and compact representations that summarize the geometric layout of lines as shown in Figure~\ref{fig:overview}.
Previous approaches for line-based localization extract rich information along the lines to establish correspondences between them~\cite{line3dpp,line_sfm_1,line_sfm_2,line_sfm_3,limap}, which often show inferior performance to prominent point features~\cite{sarlin2020superglue,lightglue} or require costly neural network inferences.
We enhance the expressive power of our descriptor exploiting relative information between lines. 
Specifically, we first define principal directions to be the representative clusters of line directions.
Then, we compute the intersections between pairs of lines having different principle directions.
The intersection points, labeled with the two principle directions they are derived from, serve as the sparse yet distinctive representation for fine-grained localization.

We further propose a fast pipeline that matches the comprehensive distribution of lines while maintaining accuracy.
We express the distribution of intersection points via point distance functions, which are the distance field measuring the closest distance to the intersection points on the spherical projection of the panorama.
To efficiently compare distance functions in various of poses, we propose a formulation that decouples rotation and translation.
Namely, we pre-compute and cache the translational variations of 3D distance functions, aligned on principal directions.
Then, during localization we quickly compare them against the rotated versions of 2D distance functions.
The matched pose can be further refined by aligning the intersection points and associated lines, as shown in Figure~\ref{fig:overview}\textcolor{red}{b}.
The resulting pipeline can quickly perform accurate localization against significantly large-scale line maps, without extensive hyperparameter tuning.

To summarize, our key contributions are as follows: (i) a fully geometric localization pipeline that overcomes the drawbacks of visual descriptors; (ii) novel spatial representations using line intersections for accurate pose search and refinement; and (iii) strategies for accelerating pose search in large-scale localization scenarios.
Due to the efficiency and robustness previously difficult to achieve by geometric methods, we expect our pipeline to serve as the practical step towards fully geometric panoramic localization.

\section{Related Work}
\label{sec:related_work}
\paragraph{Line-Based Localization}
Compared to points which are more frequently used for localization, lines can compactly describe the spatial layout of man-made structures, and thus can provide meaningful ques for localization~\cite{line_chamfer, line_transformer, sem_line, taubner2020lcd, l2d2, line_refinement,sold,gluestick,point_line_loc, ldl,limap, wglsm, hdpl, line_sfm_1, line_sfm_2, line_sfm_3, line3dpp}.
Many existing approaches for line-based localization focus on establishing one-to-one matchings between line segments in the query and the 3D map.
Here the matches are found from (i) learned features using CNNs~\cite{sold,dld}, Transformers~\cite{line_transformer} and graph neural networks~\cite{wglsm,robust_gnn_line}, or (ii) hand-crafted features describing the nearby texture~\cite{lbd,inter_context} and semantics~\cite{sem_line} of lines.
Another strand of approaches aim to jointly use points and lines during localization~\cite{gluestick,point_line_loc,limap}, which makes localization more robust in challenging scenes with repetitive structure or low texture.
For panoramic localization, LDL~\cite{ldl} is a recently proposed work that promotes distributional matching of lines via line distance functions and avoids one-to-one line matching.
Despite the effectiveness of the aforementioned approaches however, many works leverage the photometric information from the images used for line extraction, and thus are not \textit{fully geometric}.
We compare our method against exemplary line-based localization methods and demonstrate that our method can perform fast and accurate localization while only using the \textit{geometry} of lines.

\paragraph{Localization without Visual Descriptors}
While visual descriptors have been the common choice in visual localization, there has been recent works exploiting geometric cues without the aid of visual descriptors.
This is mainly motivated from (i) possible privacy breaches ~\cite{privacy_affine,privacy_analysis,privacy_concern,privacy_line2d,privacy_line3d}, and (ii) large map sizes required for storage while using visual descriptors.
Existing methods can be classified into those that propose geometric methods for pose search and other that focus on pose refinement.
For the former, LDL~\cite{ldl} and Micusik et al.~\cite{line_chamfer} exploit lines as global geometric descriptors to perform pose search.
For the latter, learning-based methods~\cite{gomatch,bpnp_net,dgcgnn} propose to describe the geometric context of keypoint locations with neural networks, and optimization-based methods~\cite{gosma,gopac} model keypoint locations as multi-modal probability distributions.
While showing competitive results, these methods still require the use of visual descriptors in the remaining localization pipelines out of their interest.
Our method can effectively localize while using geometric features for the entire pipeline, where we establish comparisons against existing methods in Section~\ref{sec:exp}.

\section{Preliminaries}
\label{sec:prelim}

Our method compares the holistic distribution of lines extracted from the 3D map and the panorama image, similar to LDL~\cite{ldl}.
In this section, we describe shared components between LDL and our approach.
Note that unlike LDL, our method only requires the line segments during localization, and thus being \textit{fully geometric} without relying on costly keypoint descriptors.

\paragraph{Input Preparation}
LDL first extracts 2D and 3D line segments using off-the-shelf line extraction methods~\cite{LSD,3d_lineseg,line3dpp}.
Throughout the remainder of the paper, we denote the 2D line segments as $L_{2D} = \{l=(s, e)\}$, where lines are represented as tuples of start and end points $s, e \in \mathbb{S}^2$ on the unit sphere.
Similarly, we denote the 3D line segments as $L_{3D}=\{\Tilde{l} = (\Tilde{s}, \Tilde{e})\}$, with the start and end points $\Tilde{s}, \Tilde{e} \in \mathbb{R}^3$.
%
LDL further extracts three principal directions for 2D and 3D, as shown in Figure~\ref{fig:pdf}\textcolor{red}{a}.
2D principal directions $D_{2D}=\{d_i\}$ are extracted by estimating the three vanishing point directions with the largest number of incident lines.
3D principal directions $D_{3D}=\{\Tilde{d}_j\}$ are similarly extracted by voting on the three most common line directions.
After discarding lines that largely deviate from the principal directions, LDL clusters lines in 2D and 3D into three sets, namely $\mathcal{L}_{2D}^{cls}=\{L_{2D}^1,L_{2D}^2,L_{2D}^3\}$ and $\mathcal{L}_{3D}^{cls}=\{L_{3D}^1,L_{3D}^2,L_{3D}^3\}$.
Each set contains the set of lines that are parallel to the designated principal direction.

\paragraph{Pose Search Using Line Distance Functions}
For each query, LDL generates a pool of $N_t \times N_r$ poses consisting of $N_t$ translations and $N_r$ rotations.
The translation pool is set as grid partitions of the 3D line map, whereas the rotation pool is set by combinatorial association between the 2D and 3D principal directions, as shown in Figure~\ref{fig:dist_func_comp}\textcolor{red}{a}.
Specifically, there can exist $2^3 \times 3!$ associations between principal directions in 2D and 3D considering sign and permutation ambiguity, where a single rotation is computed for each association using the Kabsch algorithm~\cite{kabsch}.

LDL compares line distance function of the 2D input against those generated from the pool of poses, and selects top-K poses to refine.
2D line distance functions are defined over the unit sphere $\mathbb{S}^2$ as the spherical distance to the closest line segment, namely
\begin{equation}
    f_{2D}^\text{L}(x; L_{2D}) = \min_{l \in L_{2D}} d_{\text{L}}(x, l).
    \label{eq:ldf_2d}
\end{equation}
3D line distance functions are similarly defined for each pose $R \in SO(3), t \in \mathbb{R}^3$, 
\begin{equation}
    f_{3D}^\text{L}(x; L_{3D}, R, t) = \min_{\Tilde{l} \in L_{3D}} d_{\text{L}}(x, \Pi_L(\Tilde{l}; R, t)),
    \label{eq:ldf_3d}
\end{equation}
where $\Pi_L(\Tilde{l}; R, t)$ denotes the projection of the transformed 3D line segment $(R[\Tilde{s} - t], R[\Tilde{e} - t])$ onto the unit sphere.
Based on the line distance functions, LDL then scores pose samples  $R, t$  in the pool by comparing the  function values at uniformly sampled points $Q \subset \mathbb{S}^2$ with the following cost function
\begin{equation}
    C^\text{L}(R, \! t)  = \sum_i\sum_{q \in Q} 
    \rho(f_{2D}^\text{L}(q;  L_{2D}^i) - f_{3D}^\text{L}(q;\! L_{3D}^{\sigma(i)}, R, t)),
    \label{eq:ldf_loss}
\end{equation}
where $\rho(x){=}-\mathbbm{1}\{|x| < \tau\}$ is a robust cost function and $\sigma(i)$ denotes the permutation used for estimating $R$ (i.e., $d_i \in D_{2D}$ is matched to $\Tilde{d}_{\sigma(i)} \in D_{3D}$).
Here the distance functions are separately computed for clusters of different principal directions as shown in Figure~\ref{fig:pdf}\textcolor{red}{b}, making the pose search process to better reason about the distribution of lines.

\vspace{1em}

After selecting candidate poses, LDL refines the samples of poses using conventional local feature matching~\cite{sarlin2020superglue} with PnP-RANSAC~\cite{epnp,ransac}, whereas our approach is fully geometric.
We refer the readers to the original paper~\cite{ldl} or the supplementary material for more details.
\section{Method}
As shown in Figure~\ref{fig:overview}, our localization method operates by first extracting localization inputs (Section~\ref{sec:input}), performing efficient pose search using point and line distance functions (Section~\ref{sec:pose_search}), and refining selected poses by matching line intersections (Section~\ref{sec:pose_refinement}).

\subsection{Input Preparation}
\label{sec:input}
\begin{figure}[t]
  \centering
    \includegraphics[width=\linewidth]{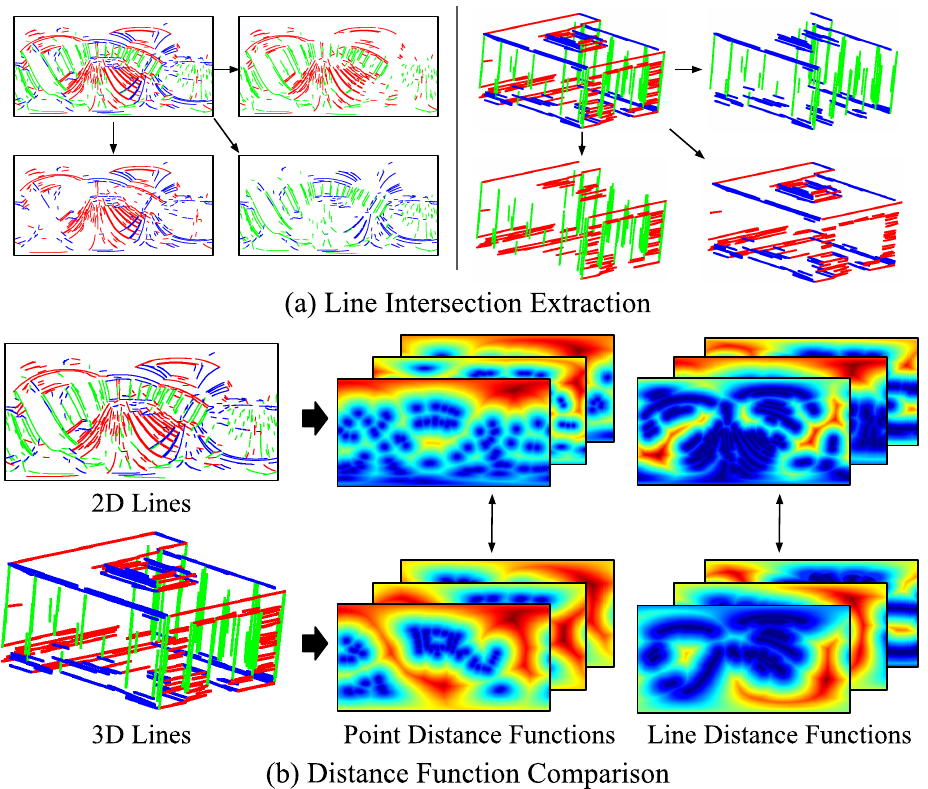}
   \caption{Line intersection extraction and distance function comparison. (a) We pairwise intersect lines clustered along the principal directions and obtain three groups of intersection points. (b) The intersection point clusters are used to define point distance functions. Together with line distance functions that describe the coarse scene layout, point distance functions describe the fine-grained geometry of lines which enable accurate pose search.}
   \label{fig:pdf}
\vspace{-1em}
\end{figure}

In addition to lines and principal directions, we introduce \textit{line intersection points} as input to enhance localization performance.
As shown in Figure~\ref{fig:pdf}\textcolor{red}{a}, 
we obtain three clusters of line intersection points for both 2D and 3D, namely $\mathcal{P}_{2D}^{cls}=\{P_{2D}^{12}, P_{2D}^{23}, P_{2D}^{31}\}$ and $\mathcal{P}_{3D}^{cls}=\{P_{3D}^{12}, P_{3D}^{23}, P_{3D}^{31}\}$.
Here $P_{2D}^{ij}$ is the set of line intersections obtained from $L_{2D}^i$ and $L_{2D}^j$, and $P_{3D}^{ij}$ is similarly defined.
Additional details about input preparation are deferred to the supplementary material.

\subsection{Pose Search}
\label{sec:pose_search}


\subsubsection{Point Distance Functions}
We use the additional input of line intersection points to devise point distance functions.
As point distance functions contain more fine-grained information of scene keypoints, they can guide towards making more accurate pose search. 

Point distance functions are defined in a similar way as line distance functions.
Given a set of 2D points $P_{2D}$, 2D point distance functions are defined over the unit sphere $\mathbb{S}^2$ as the distance to the closest point in $P_{2D}$, namely
\begin{equation}
    f_{2D}^\text{P}(x; P_{2D}) = \min_{p \in P_{2D}} d_{\text{P}}^\gamma (x, p),
    \label{eq:pdf_2d}
\end{equation}
where $d_{\text{P}}(x, p) = \cos^{-1} \langle x, p \rangle$ and $\gamma = 0.2$ is a sharpening parameter to incur more dramatic change near the point locations. 
3D point distance functions are defined similarly for rotations and translations within the 3D map,
\begin{equation}
    f_{3D}^\text{P}(x; P_{3D}, R, t) = \min_{\Tilde{p} \in P_{3D}} d^\gamma_{\text{P}}(x, \Pi(\Tilde{p}; R, t)),
    \label{eq:pdf_3d}
\end{equation}
where $\Pi(\Tilde{p}; R, t)$ denotes the spherical projection of the transformed 3D point $R[\Tilde{p} - t]$.
As in Equation~\ref{eq:ldf_loss}, point distance functions can be separately compared for intersection point clusters as follows,
\begin{equation}
    C^\text{P}(R, \! t) = \sum_{i\neq j}\sum_{q \in Q} 
    \rho(f_{2D}^\text{P}(q; \! P_{2D}^{ij}) - f_{3D}^\text{P}(q;\! P_{3D}^{\sigma(ij)},\! R,\! t)),
    \label{eq:pdf_loss}
\end{equation}
where with an abuse of notation $\sigma(ij)$ is the juxtaposition of $\sigma(i)$ and $\sigma(j)$, namely permutations of principal directions for estimating $R$.
As explained in Section~\ref{sec:prelim} and Figure~\ref{fig:dist_func_comp}\textcolor{red}{a}, each rotation in the rotation pool is obtained by combinatorially associating principal directions.
Our method then uses point distance functions \textit{along with} line distance functions for pose search, 
\begin{equation}
    C(R,t) =  C^\text{L}(R, t) + C^\text{P}(R, t). \label{eq:cost_sum}
\end{equation}

\subsubsection{Efficient Distance Function Comparison}
\begin{figure}[t]
  \centering
    \includegraphics[width=\linewidth,height=300pt]{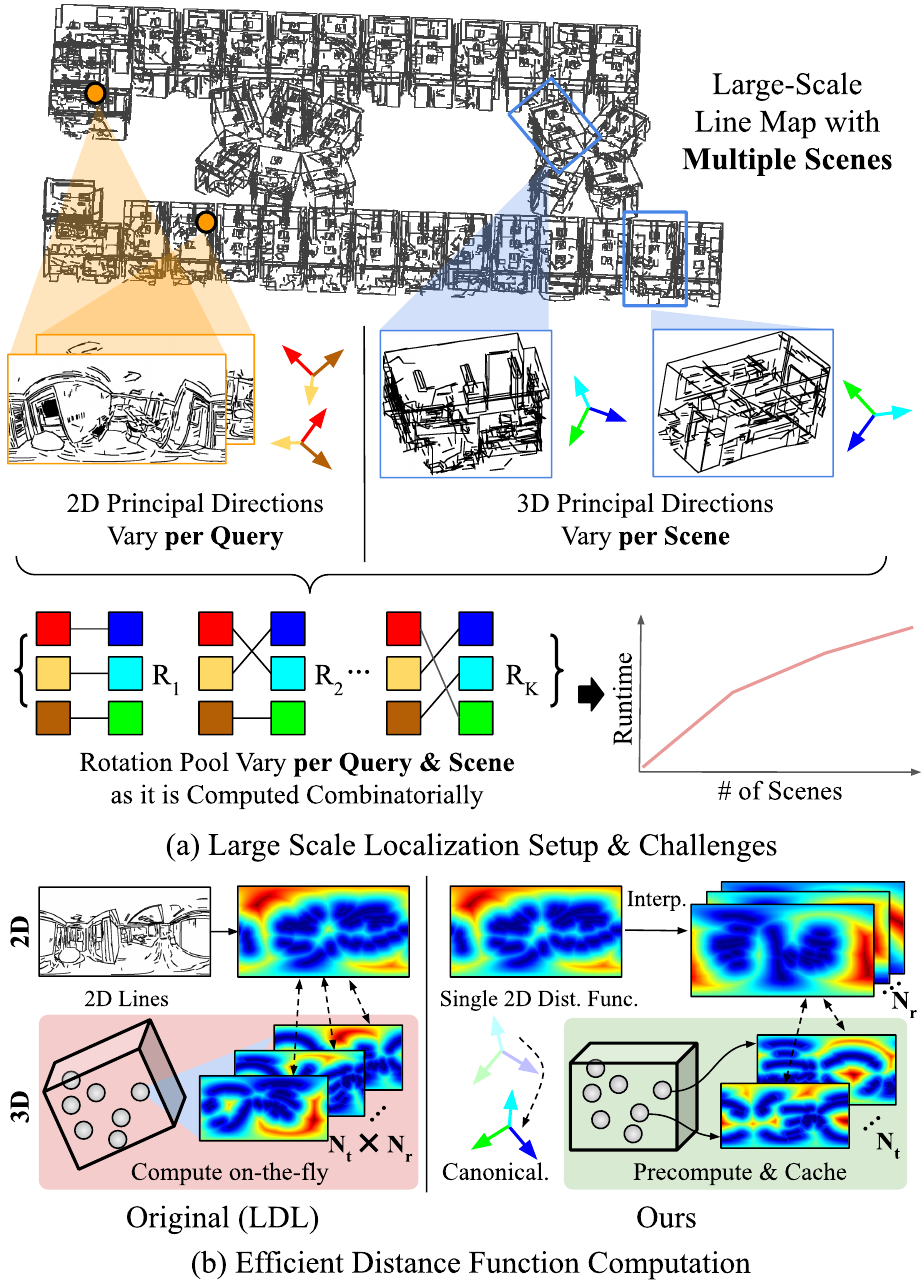}
   \caption{Motivation and overview of efficient distance function comparison. (a) In large-scale localization scenarios, the rotation pool constantly changes due to the variability of principal directions in 2D and 3D. Thus exhaustively computing 3D distance functions for all possible poses on the fly leads to large runtime. (b) We instead propose to (i) decouple translation and rotation, (ii) precompute and cache 3D distance functions aligned in the canonical direction, and (iii) during localization interpolate 2D distance function values at various rotations greatly reducing computation.
   }
   \label{fig:dist_func_comp}
\vspace{-1em}
\end{figure}

Given a pool of $N_t \times N_r$ poses, our method selects top-$K$ poses to refine by comparing Equation~\ref{eq:cost_sum}.
However, exhaustive comparison cannot scale to large scenes containing multiple line maps (or rooms) as shown in Figure~\ref{fig:dist_func_comp}\textcolor{red}{a}.
The 3D distance function term depends on the rotation pool, which changes for different query images as their 2D principal directions change.
Therefore the 3D distance functions have to be evaluated \textit{on-the-fly} for each query image for all line maps in 3D.
Below we introduce two strategies for efficient pose search that is scalable to large scale localization.

\paragraph{Pre-computing 3D Distance Functions}
We observe that rotation from 3D distance functions can be decoupled so that the distance functions only for different translations can be pre-computed and \textit{cached.}
%
Specifically, we propose the following modified cost functions,
\begin{align}
    \label{eq:mod_ldf_cost}
    C^\text{L}(R, \! t) &\! = \! \sum_i\sum_{q \in Q} \rho(f_{2D}^\text{L}(q;  R^TL_{2D}^i) - \underbrace{f_{3D}^\text{L}(q;\! L_{3D}^{\sigma(i)}, I, t))}_{\text{pre-compute \& cached}}, \\
    \label{eq:mod_pdf_cost}
    C^\text{P}(R, \! t) &\! = \! \sum_{i\neq j}\sum_{q \in Q} \rho(f_{2D}^\text{P}(q; \! R^TP_{2D}^{ij}) - \underbrace{f_{3D}^\text{P}(q;\! P_{3D}^{\sigma(ij)},\! I,\! t))}_{\text{pre-compute \& cached}},
\end{align}
where $R^TL^i_{2D}, R^TP^{ij}_{2D}$ denotes 2D lines and points rotated by $R^T$ respectively.


The proposed formulation indicates applying $N_r$ rotations on 2D distance functions and comparing against the cached 3D distance functions.
When pre-computing 3D distance functions, we propose to align the 3D principal directions to a common coordinate frame (i.e., \textit{canonicalize} 3D lines), as shown in Figure~\ref{fig:dist_func_comp}\textcolor{red}{b}.
This alleviates calculating all possible rotations associating 2D-3D principal directions, but focus only on 2D rotations relative to the canonicalized 3D principal directions.
Therefore the 2D line distance functions can be computed in a constant runtime, independent to the number of line maps.

\paragraph{Fast 2D Distance Functions via Interpolation}
We further introduce a method to accelerate 2D distance function extraction.
Specifically, we evaluate distance function only once at the identity rotation, and use them to estimate values at $N_r$ rotations as shown in Figure~\ref{fig:dist_func_comp}\textcolor{red}{b}.
Since distance functions are defined on a sphere, rotation of the function is equivalent to rotating the query points $q \in Q$ as
\begin{equation}
    f_{2D}(q; \hat{R}L_{2D}) = f_{2D}(\hat{R}^Tq; L_{2D}) \approx f_{2D}(q^*;L_{2D}),
\label{eq:interp_2d}
\end{equation}
where $q^* = \argmin_{\hat{q}\in Q} d_\text{P}(q, \hat{R}^T\hat{q})$, or the nearest neighbor interpolation.

While the new cost functions and the interpolation scheme can effectively reduce runtime for pose search, they are \textit{approximations} of the original cost functions.
Nevertheless, we find that the approximations are sufficiently accurate, due to the following theorem.

\paragraph{Theorem 1.}
\textit{
Given a metric $d(\cdot, \cdot)$ defined over the unit sphere $\mathbb{S}^2$, let $f(x; S){:=} \min_{s \in S} d(x, s)$ denote a distance function to a set of spherical points $S \! \subset \! \mathbb{S}^2$.
Consider a countable, finite set of spherical points $Q \! \subset \! \mathbb{S}^2$ that satisfy $\max_{q \in Q}\min_{\hat{q} \in Q} d(q, R\hat{q}) \! \leq \! \max_{q \in Q}\min_{\hat{q} \neq q} d(q, \hat{q}) \! = \! \delta$ for all $R \in SO(3)$.
For an arbitrary rotation $\tilde{R} \in SO(3)$, the following inequality holds:
}
\begin{equation}
\frac{1}{|Q|}\sum_{q \in Q} |f(q; S) - f(\argmin_{\hat{q}\in Q} d(\hat{q}, \tilde{R}q); \tilde{R}S)|\leq \delta.
\label{eq:approx_bound}
\end{equation}
We provide detailed proofs to the theorem in the supplementary material.
The theorem suggests that for \textit{sufficiently dense} query points, the sum of distance function values remain almost constant under rotation, namely $\sum_{q\in Q}f(q;S) \approx \sum_{q\in Q}f(q;RS)$.
As a result, we can substitute the lines and points in the original cost functions with their rotated counterparts: for example for lines,
\begin{align}
    &\sum_i\sum_{q \in Q} \rho(f_{2D}^\text{L}(q;  L_{2D}^i) - f_{3D}^\text{L}(q;\! L_{3D}^{\sigma(i)}, R, t)) \\
    &\approx \sum_i\sum_{q \in Q} \rho(f_{2D}^\text{L}(q;  R^TL_{2D}^i) - f_{3D}^\text{L}(q;\! L_{3D}^{\sigma(i)}, I, t)),
\end{align}
which leads to the modified cost function in Equation~\ref{eq:mod_ldf_cost},~\ref{eq:mod_pdf_cost}.

\subsection{Pose Refinement}
\label{sec:pose_refinement}

After finding the top-$K$ poses $\{(R_k, t_k)\}$, we can refine them only using geometric representations.
Given a good initial pose, we establish correspondences in 2D and 3D from nearest neighbor matching, and refine translation then rotation to find a highly accurate pose.

\begin{figure}[t]
  \centering
    \includegraphics[width=\linewidth]{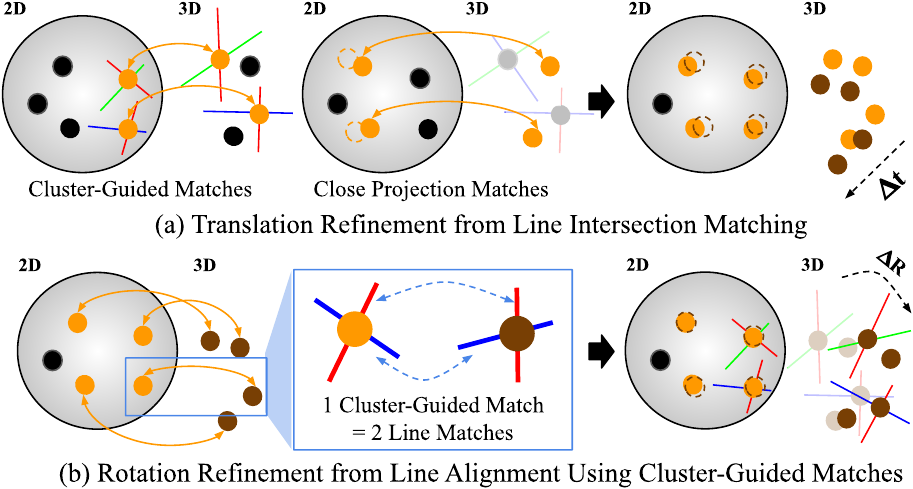}
   \caption{Pose refinement using line intersections. (a) We first match line intersections that belong to the same cluster type using mutual nearest neighbors (Cluster-Guided Matches), along with a small pair of 2D, 3D intersections that are sufficiently close together on the sphere (Close Projection Matches). The matches are then used to optimize translation, where matches are also iteratively updated similar to ICP~\cite{icp}. (b) Rotation is then optimized based on the final set of matches by aligning the incident line directions of cluster-guided matches. Here we exploit the fact that each cluster-guided match yields a pair of line matches.}
   \label{fig:refine}
\vspace{-1em}
\end{figure}

\paragraph{Line Intersection Matching}
Our refinement, in a big picture, is similar to the iterative closest point (ICP)~\cite{icp}, where
fast and accurate correspondences are crucial.
As shown in Figure~\ref{fig:refine}\textcolor{red}{a}, we avoid outliers by considering two types of matches, namely intersection points with (i) coherent principal direction clusters, or (ii) sufficiently close projected distances.
For the former \textit{cluster-guided matches}, we use the permutation $\sigma(ij)$ for estimating $R_k$ of the candidate pose.
We find the mutual nearest neighbor~\cite{mutual_nn} between the corresponding intersection cluster pairs: $P_{2D}^{ij} \subset \mathcal{P}^{cls}_{2D}$ and $P_{3D}^{\sigma(ij)} \subset \mathcal{P}^{cls}_{3D}$.
Specifically, we first project 3D intersections onto the sphere $\Pi(R_k[P_{3D}^{\sigma(ij)} - t_k])$ and find matches with $P_{2D}^{ij}$.
For the latter \textit{close projection matches}, we ignore the cluster types and project all 3D intersections onto the sphere. 
We then retrieve 2D-3D pairs with distances below a threshold $\delta{=}0.1$, resulting in an initial set of point matches $\mathcal{M}^\text{P} = \{(m_{2D}, m_{3D})\}$.

\paragraph{Translation Refinement}
We fix rotation estimated from principal directions in Section~\ref{sec:prelim} and only optimize for translation given the set of point matches.
Specifically, we minimize the following cost function with respect to translation $t_k$ using gradient descent~\cite{adam, sgd}:
\begin{equation}
    C^\text{trans}(t_k^{(n)}) = \sum_{m \in \mathcal{M}^\text{P}} \|m_{2D} - \Pi(R_k[m_{3D} - t_k^{(n)}])\|_1,
\label{eq:trans_refine}
\end{equation}
where $t_k^{(n)}$ is the translation at step $n$.
After each step we also update the set of matches $\mathcal{M}^\text{P}$ via nearest neighbor search between $P_{2D}^{ij}$ and $\Pi(R_k[P_{3D}^{\sigma(ij)} - t_k^{(n)}])$.
The refined translation $\hat{t}_{k^*}$ with the smallest cost value and its associated matches $\widehat{\mathcal{M}}^\text{P}$ are then passed to rotation refinement.

\paragraph{Rotation Refinement}
Finally, we deviate from intersection points, and refine the rotation using the original line directions.
As shown in Figure~\ref{fig:refine}\textcolor{red}{b}, we can deduce two pairs of  \textit{line matches} from each cluster-guided match in $\widehat{\mathcal{M}}^\text{P}$.
Let $\widehat{M^\text{L}}=\{([s_{2D}, e_{2D}], [\tilde{s}_{3D}, \tilde{e}_{3D}])\}$ denote the set of line matches obtained from $\widehat{\mathcal{M}}^\text{P}$.
We iteratively minimize the following cost function with respect to rotation $R_{k^*}$,
\begin{equation}
    C^\text{rot}(R_{k^*}^{(n)}) =\sum_{m \in \widehat{M^\text{L}}} |\cos (\frac{s_{2D} \times e_{2D}}{\|s_{2D} \times e_{2D}\|}, \frac{R_{k^*}^{(n)} [\tilde{s}_{3D} - \tilde{e}_{3D}]}{\|\tilde{s}_{3D} - \tilde{e}_{3D}\|})|.
\label{eq:rot_refine}
\end{equation}
The cost function aligns the line directions in 2D to the rotated line directions in 3D.
Combined with efficient pose search, our sequential pose refinement scheme enables accurate pose estimation without resorting to feature descriptors, where we perform detailed comparisons against feature-based methods in Section~\ref{sec:exp}.

\vspace{-0.5em}
\section{Experiments}
\label{sec:exp}
\vspace{-0.5em}

In this section, we evaluate our method on a wide range of localization scenarios.
We use two commonly used datasets~\cite{piccolo, cpo, ldl, sphere_cnn, gosma}: Stanford 2D-3D-S~\cite{stanford2d3d} and OmniScenes~\cite{piccolo}.
Stanford 2D-3D-S consists of 1413 panoramas captured in 272 rooms, while OmniScenes is a recently proposed dataset that consists of 4121 panoramas from 7 rooms.
Unless specified otherwise, we use the entire Stanford 2D-3D-S dataset and the Extreme split from OmniScenes, which is the most challenging split that contains both scene changes and fast camera motion.
For all of scenarios, we use the fixed set of hyperparameters with the inlier threshold $\tau=0.1$, number of query points $|Q|=642$, and number of translations per line map $N_t=500$.
Our method is implemented in PyTorch~\cite{pytorch}, and we use a single RTX2080 GPU with an Intel Core i7-7500U CPU.

\begin{table}[t]
\centering
\resizebox{\linewidth}{!}{

\begin{tabular}{l|
cccc|ccc}
\toprule
Scene No. & 1 & 2 & 3 & 4 & 
\multirow{2}{*}[-2.5pt]{\begin{tabular}[c]{@{}c@{}} Search\\Time (s)\end{tabular}}
& \multirow{2}{*}[-2.5pt]{\begin{tabular}[c]{@{}c@{}}Refine\\Time (s)\end{tabular}} 
& \multirow{2}{*}[-2.5pt]{\begin{tabular}[c]{@{}c@{}}Map\\Size (GB)\end{tabular}} \\
\cmidrule{1-5}
\# of Rooms & 7 & 44 & 40 & 60 &  &  &  \\
\midrule
LDL & 0.54 & 0.26 & 0.21 & 0.19 & 0.24 & 0.07 & 129.75 \\
CP + LG & 0.66 & 0.58 & 0.67 & \textbf{0.69} & 0.23 & \textbf{0.07} & 131.11 \\
CP + GS & 0.61 & 0.58 & \textbf{0.73} & 0.61 & 0.23 & 0.21 & 132.76 \\
CP + GM & 0.25 & 0.27 & 0.35 & 0.33 & 0.23 & 0.53 & 2.86 \\
SFRS + LG & 0.60 & \textbf{0.61} & 0.66 & 0.68 & 0.10 & 0.07 & 132.47 \\
SFRS + GS & 0.47 & 0.58 & 0.60 & 0.58 & 0.10 & 0.21 & 134.12 \\
SFRS + GM & 0.07 & 0.23 & 0.31 & 0.23 & 0.10 & 0.53 & 4.22 \\
Ours & \textbf{0.76} & \textbf{0.61} & 0.68 & 0.64 & \textbf{0.02} & 0.38 & \textbf{2.56} \\
\bottomrule
\end{tabular}
}
\caption{Localization evaluation results in large scale scenes. We compare our method against LDL~\cite{ldl} along with combinations of various pose search (Cosplace~\cite{cosplace} (CP), SFRS~\cite{openibl}) and refinement (LightGlue~\cite{lightglue} (LG), Gluestick (GS), GoMatch (GM)) methods. For each scene, we report the localization accuracy at 0.1m and $5^\circ$. We additionally report the average pose search time, refinement time, and map size during localization.}
\label{table:large_scale}
\vspace{-1em}
\end{table}
\subsection{Large-scale Localization}
We first provide evaluation in large-scale localization scenarios, where the geometric methods as ours can benefit from the lightweight  representations.
Here we join multiple room-level line maps in OmniScenes~\cite{piccolo} and Stanford 2D-3D-S~\cite{stanford2d3d} to create four large-scale maps: \textit{split 1 }containing all rooms in OmniScenes, \textit{split 2} containing all rooms in Area 1 from Stanford 2D-3D-S, \textit{split 3} and \textit{split 4} containing 40 and 60 office rooms in Stanford 2D-3D-S respectively.
Note we design the latter two splits to examine localization in large number of similar-looking structures.
An exemplary large scale map is shown in Figure~\ref{fig:overview}.

In Table~\ref{table:large_scale} we compare against LDL~\cite{ldl} and competitive combinations of neural network-based pose search and refinement methods.
Note that for baselines other than LDL, we synthesize images by projecting the colored point cloud to extract descriptors, similar to ~\cite{cpo, ldl}.
Here all methods extract top-5 retrieval from pose search, and refine them.
Our method largely outperforms LDL and demonstrates competitive performance against neural network-based methods.
Further, due to the efficient pose search scheme proposed in Section~\ref{sec:pose_search}, our method exhibits an order-of-magnitude shorter search time compared to the baselines.
In addition, as our method only stores the $|Q|$ 3D distance function values, the map size is much smaller than the baselines that cache high-dimensional global/local feature descriptors.
We make a detailed analysis of the map size in the supplementary material.

\subsection{Pose Search}
\label{sec:pose_search}

\begin{figure}[t]
  \centering
    \includegraphics[width=\linewidth]{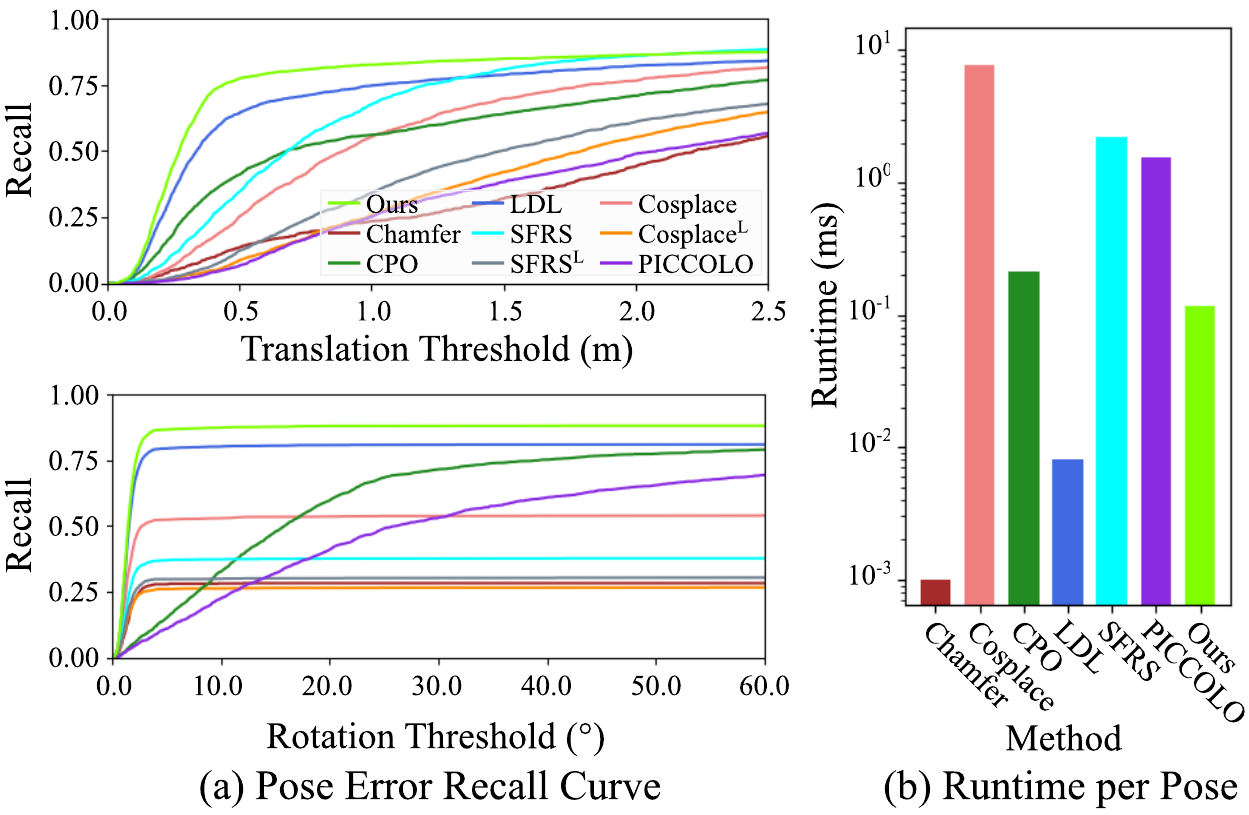}
   \caption{Pose error recall and runtime comparison in OmniScenes~\cite{piccolo} and Stanford 2D-3D-S~\cite{stanford2d3d} for the top-1 retrieval results. Note the runtime is plotted in log scale, and the superscript X\textsuperscript{L} denotes that the baseline takes line images as input.}
   \label{fig:pose_search}
   \vspace{-1.5em}
\end{figure}

We make further analysis on the first stage, i.e., pose search, where the task is to quickly find the closest pose from pools of rotations and translations in a 3D map.
The compared baselines are categorized into neural network-based (SFRS~\cite{openibl}, Cosplace~\cite{cosplace}); color distribution-based (PICCOLO~\cite{piccolo}, CPO~\cite{cpo}); and line-based (LDL~\cite{ldl}, Chamfer~\cite{line_chamfer}) methods.
We use the colored point cloud from the tested datasets and render synthetic views at various translations and rotations, from which we prepare neural network descriptors for the baselines.
For fair comparison, we use the identical pool of translations and rotations.
We additionally implement neural network-based methods that take line images as input instead of the original RGB images (SFRS\textsuperscript{L}, Cosplace\textsuperscript{L}), to examine how these methods can handle geometric localization scenarios.

Figure~\ref{fig:pose_search} shows the top-1 recall curves and runtimes for generating a global descriptor per pose.
Our method attains a high recall even at strict translation and rotation thresholds, and shows a much shorter runtime than neural network-based or color-based methods.
While several line-based methods (LDL, Chamfer) exhibit a shorter runtime, the retrieved poses from there methods are relatively inaccurate compared to our method.

\vspace{-1em}
\begin{table}[t]
    \begin{subtable}{\linewidth}
    \centering
    \resizebox{0.65\linewidth}{!}{
    \begin{tabular}{lccc}
    \toprule
    Method & \begin{tabular}[c]{@{}c@{}}Accuracy\\ (0.1m, $5^\circ$)\end{tabular} & \begin{tabular}[c]{@{}c@{}}Accuracy\\ (0.2m, $10^\circ$)\end{tabular} & \begin{tabular}[c]{@{}c@{}}Accuracy\\ (0.3m, $15^\circ$)\end{tabular} \\
    \midrule
    LDF & 0.57 & 0.58 & 0.58 \\
    LDF + PDF & \textbf{0.61} & \textbf{0.62} & \textbf{0.63} \\
    \bottomrule
    \end{tabular}
    }
    \caption{Pose search using line / point distance functions (LDF / PDFs)}
    \end{subtable}

    \begin{subtable}{\linewidth}
    \centering
    \resizebox{0.78\linewidth}{!}{
    \begin{tabular}{lccc}
    \toprule
    Method & \begin{tabular}[c]{@{}c@{}}Accuracy\\ (0.1m, $5^\circ$)\end{tabular} & \begin{tabular}[c]{@{}c@{}}Runtime (s)\\ @ GPU\end{tabular} & \begin{tabular}[c]{@{}c@{}}Runtime (s)\\@ CPU\end{tabular} \\
    \midrule
    Ours w/o Decoupling & \textbf{0.62} & 45.049 & 1908.522 \\
    Ours & 0.61 & \textbf{2.890} & \textbf{109.561} \\
    \bottomrule
    \end{tabular}
    }
    \caption{Decoupling rotation and translation for 3D distance functions}
    \end{subtable}

    \begin{subtable}{\linewidth}
    \centering
    \resizebox{0.75\linewidth}{!}{
    \begin{tabular}{ccccc}
    \toprule
    \multicolumn{1}{l}{Interp.} & \multicolumn{1}{l}{Canonical.} & \begin{tabular}[c]{@{}c@{}}Accuracy\\ (0.1m, $5^\circ$)\end{tabular} & \begin{tabular}[c]{@{}c@{}}Runtime (s)\\ @ GPU\end{tabular} & \begin{tabular}[c]{@{}c@{}}Runtime (s)\\@ CPU\end{tabular} \\
    \midrule
    \xmark & \xmark & \textbf{0.61} & 0.396 & 7.243 \\
    \xmark & \cmark & \textbf{0.61} & 0.007 & 0.198 \\
    \cmark & \cmark & \textbf{0.61} & \textbf{0.004} & \textbf{0.080} \\
    \bottomrule
    \end{tabular}
    }
    \caption{Interpolation and line canonicalization for 2D distance functions}
    \end{subtable}
    \caption{Ablation study of our pose search method, using the entire Area 1 of Stanford 2D-3D-S as the 3D map. Note the runtime for distance functions indicates the time needed to generate \textit{all} 2D or 3D distance functions for rooms in the map.}
    \label{table:ablation_search}
    \vspace{-1.5em}
\end{table}

\paragraph{Ablation Study}
We ablate the key constituents of the fast and accurate pose search pipeline, namely the point distance functions and the efficient distance function computation.
Table~\ref{table:ablation_search} reports the localization error and runtime for top-5 retrieval using the entire Area 1 of Stanford 2D-3D-S~\cite{stanford2d3d}.
Table~\ref{table:ablation_search}\textcolor{red}{a} suggests that using point distance functions along with line distance functions enhance localization performance.
While the LDF-only method is similar to LDL, we can still observe a performance gap compared to Table~\ref{table:large_scale}.
Due to the efficient distance function comparison, our method can seamlessly handle large number ($|Q|=642$) of query points during pose search, whereas LDL can only handle a limited number of query points ($|Q|=42$).
We provide additional experiments regarding this aspect in the supplementary material.
Table~\ref{table:ablation_search}\textcolor{red}{b} and~\ref{table:ablation_search}\textcolor{red}{c} further support this claim: decoupling rotation/translation for 3D distance functions and interpolating and canonicalizing 2D distance functions lead to significant runtime drops, while showing almost no loss in accuracy.

\subsection{Pose Refinement}

\begin{table}[t]
\centering
\resizebox{\linewidth}{!}{


\begin{tabular}{llc|ccc}
\toprule
\begin{tabular}[c]{@{}l@{}}Refinement \\ Type\end{tabular} & Method & \multicolumn{1}{l|}{\begin{tabular}[c]{@{}l@{}}Visual\\ Desc.\end{tabular}} & \begin{tabular}[c]{@{}c@{}}Accuracy \\ (0.1m, $5^\circ$)\end{tabular} & \begin{tabular}[c]{@{}c@{}}Accuracy \\ (0.2m, $10^\circ$)\end{tabular} & \begin{tabular}[c]{@{}c@{}}Accuracy \\ (0.3m, $15^\circ$)\end{tabular} \\
\midrule
\multirow{2}{*}{\begin{tabular}[c]{@{}l@{}}Line-Based \\ Refinement\end{tabular}} & Line Transformer~\cite{line_transformer} & $\bigcirc$ & 0.76 & 0.79 & 0.80 \\
 & Gluestick~\cite{gluestick} & $\bigcirc$ & 0.75 & 0.79 & 0.80 \\
\midrule
\multirow{5}{*}{\begin{tabular}[c]{@{}l@{}}Point-Based \\ Refinement\end{tabular}} & SuperGlue~\cite{sarlin2020superglue} & $\bigcirc$ & 0.78 & \textbf{0.80} & \textbf{0.81} \\
 & LightGlue~\cite{lightglue} & $\bigcirc$ & \textbf{0.79} & \textbf{0.80} & \textbf{0.81} \\
 & LoFTR~\cite{loftr} & $\bigcirc$ & 0.74 & 0.78 & 0.78 \\
 & PICCOLO~\cite{piccolo} & $\bigcirc$ & 0.57 & 0.59 & 0.60 \\
 & CPO~\cite{cpo} & $\bigcirc$ & 0.62 & 0.64 & 0.65 \\
 \midrule
\multirow{10}{*}{\begin{tabular}[c]{@{}l@{}}Geometric \\ Refinement\end{tabular}} & GoMatch~\cite{gomatch} & \xmark & 0.61 & 0.72 & 0.74 \\
 & BPnPNet~\cite{bpnp_net} & \xmark & 0.04 & 0.22 & 0.41 \\
 & Gao et. al ~\cite{line_refinement} & \xmark & 0.21 & 0.56 & 0.66 \\
 & PDF Minimization & \xmark & 0.22 & 0.38 & 0.47 \\
 & LoFTR\textsuperscript{L}~\cite{loftr} & \xmark & 0.09 & 0.31 & 0.50 \\
 & SuperGlue\textsuperscript{L}~\cite{sarlin2020superglue} & \xmark & 0.06 & 0.34 & 0.57 \\
 & LightGlue\textsuperscript{L}~\cite{lightglue} & \xmark & 0.37 & 0.57 & 0.67 \\
 & GlueStick\textsuperscript{L}~\cite{gluestick} & \xmark & 0.32 & 0.51 & 0.62 \\
 & Line Transformer\textsuperscript{L}~\cite{line_transformer} & \xmark & 0.06 & 0.29 & 0.49 \\
 & Ours & \xmark & 0.67 & 0.74 & 0.76\\
 \bottomrule
\end{tabular}

}
\vspace{-0.5em}
\caption{Pose refinement evaluation in OmniScenes~\cite{piccolo} and Stanford 2D-3D-S~\cite{stanford2d3d}. Note the superscript X\textsuperscript{L} denotes that the baseline takes line images as input.}
\label{table:pose_refine}
\vspace{-1.em}
\end{table}

To evaluate the second stage of pose refinement,  we retrieve the same top-1 pose using our pose search scheme, and perform refinement using various methods.
We compare the accuracy against three classes of refinement methods: line-based~\cite{line_transformer, gluestick}, point-based~\cite{sarlin2020superglue,lightglue,loftr,piccolo,cpo}, and geometric~\cite{gomatch,bpnp_net,line_refinement}.
While line and point-based methods find matches using the photometric information near the lines and points to increase accuracy, geometric methods only rely on the locations of keypoints (GoMatch~\cite{gomatch}, BPnPNet~\cite{bpnp_net}) or lines, as we propose.
We also implement the line alignment method proposed from Gao et al.~\cite{line_refinement}, along with a conceived baseline that directly minimizes Equation~\ref{eq:pdf_loss} with gradient descent (PDF minimization).

Table~\ref{table:pose_refine} shows that our method consistently outperforms all the tested geometric methods, while showing competitive performance against methods that use visual descriptors.
Considering the minimal map size (Table~\ref{table:large_scale}), it is a noticeable performance improvement of purely geometric approaches.

\begin{table}[]
\centering
\resizebox{0.8\linewidth}{!}{
\begin{tabular}{lccc}
\toprule
Method & \textit{t}-error & \textit{R}-error & \begin{tabular}[c]{@{}c@{}}Accuracy\\ (0.1m, $5^\circ$)\end{tabular} \\
\midrule
Only trans. refine & \textbf{0.06} & 1.56 & \textbf{0.78} \\
No intersection clusters & 0.08 & 1.11 & 0.63 \\
Ours & \textbf{0.06} & \textbf{1.05} & 0.77 \\
\bottomrule
\end{tabular}
}
\vspace{-0.5em}
\caption{Ablation study of key components of our pose refinement method, using the Room 2 subset from OmniScenes~\cite{piccolo}.}
\label{table:abl_refine}
\vspace{-1em}
\end{table}

\paragraph{Ablation Study}
Our pose refinement scheme deploys line intersections in the place of conventional visual descriptors and effectively disambiguates matches from point clusters.
Table~\ref{table:abl_refine} displays the effect of those components in terms of localization accuracy and the median translation/rotation errors using Room 2 from OmniScenes~\cite{piccolo}.
Optimizing only translation leads to larger rotation errors, which suggests the importance of rotation refinement.
In addition, directly performing nearest neighbor matching without using intersection clusters lead to a large drop in localization performance.
As intersection clusters prevent outlier matches, they are crucial for accurate pose refinement.
\vspace{-0.5em}


\begin{table}[t]
\resizebox{\linewidth}{!}{
\begin{tabular}{lc|cccc|cc}
\toprule
Method & \multicolumn{1}{l|}{\begin{tabular}[c]{@{}l@{}}Visual \\ Desc.\end{tabular}} & Orig. & Intensity & Gamma & \begin{tabular}[c]{@{}c@{}}White \\ Balance\end{tabular} & Range & Std \\
\midrule
Line Transformer & $\bigcirc$ & 0.88 & 0.70 & 0.86 & 0.89 & 0.22 & 0.09 \\
Gluestick & $\bigcirc$ & \textbf{0.89} & \textbf{0.80} & \textbf{0.89} & \textbf{0.90} & 0.12 & 0.05 \\
\midrule
GoMatch & \xmark & 0.67 & 0.66 & 0.63 & 0.67 & 0.08 & 0.03 \\
Gluestick\textsuperscript{L} & \xmark & 0.36 & 0.49 & 0.37 & 0.31 & 0.22 & 0.09 \\
Ours & \xmark & 0.77 & 0.74 & 0.74 & 0.77 & \textbf{0.05} & \textbf{0.02} \\
\bottomrule
\end{tabular}
}
\vspace{-0.5em}
\caption{Pose refinement evaluation under varying lighting conditions in OmniScenes~\cite{piccolo}. We report the localization accuracy at 0.1m and $5^\circ$, along with their range and standard deviations.}
\label{table:lighting_refine}
\vspace{-0.5em}
\end{table}

\subsection{Robustness Evaluation}
\vspace{-0.5em}
\paragraph{Illumination Changes} Our method can stably perform localization in various environment conditions as long as we can extract lines. 
We demonstrate the robustness of our method amidst lighting changes using the OmniScenes~\cite{piccolo} dataset.
Specifically, we separately measure the performance of pose search and refinement after applying synthetic color variations to the input panorama, similar to the experiments conducted in LDL~\cite{ldl}.
We consider three types of color variations (intensity, gamma, and white balance), where we apply two levels of variations for each type.

\begin{figure}[t]
  \centering
    \includegraphics[width=\linewidth]{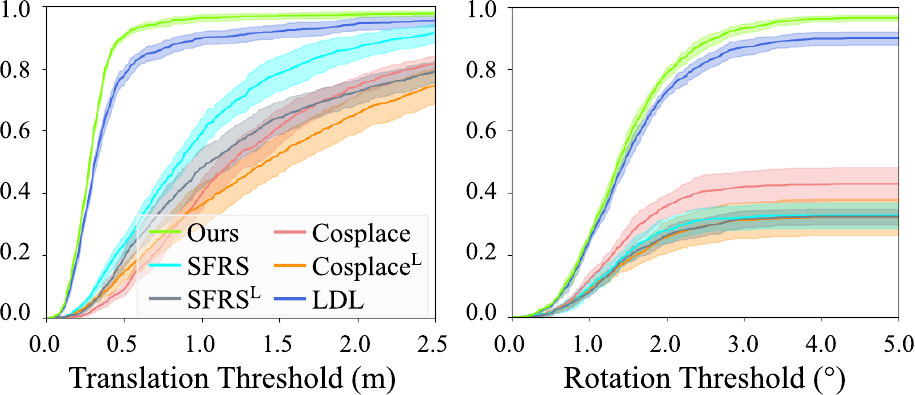}
   \caption{Recall curves of top-1 pose search amidst lighting condition variations. We evaluate localization performance on six lighting variations and shade the maximum and minimum recall. The solid lines are the averaged recall values for each tested baseline.}
   \label{fig:lighting_pose_search}
\vspace{-1.5em}
\end{figure}

Table~\ref{table:lighting_refine} displays the pose refinement performance at varying lighting conditions, where all the tested methods take the top-1 pose search results using our method as input.
Our refinement scheme performs competitively against visual descriptor-based methods, while showing a much smaller range and standard deviations in localization accuracy.
A similar trend is observable in Figure~\ref{fig:lighting_pose_search} that shows the recall curves of top-1 retrievals from various pose search methods.
The width of the shaded regions, which indicates the range of recall values under illumination changes, is much smaller for our method compared to the baselines.
By using lines and their intersections as the only cues for localization, our method remains robust against lighting changes.

\vspace{-1.5em}

\paragraph{Applicability to Floorplan Localization}
\begin{table}[]
\resizebox{\linewidth}{!}{
\begin{tabular}{lll|cc}
\toprule
Method & \begin{tabular}[c]{@{}l@{}}Additional\\ Input\end{tabular} & \begin{tabular}[c]{@{}l@{}}Estimation\\ Target\end{tabular} & \begin{tabular}[c]{@{}c@{}}Accuracy\\ (0.1m, $5^\circ$)\end{tabular} & \begin{tabular}[c]{@{}c@{}}Accuracy\\ (1m, $30^\circ$)\end{tabular} \\
\midrule
LaLaLoc~\cite{lalaloc} & Layout Depth & 2D Trans. & 0.91 & 0.95 \\
LaLaLoc~\cite{lalaloc} & N/A & 2D Trans. & 0.58 & 0.88 \\
LaLaLoc++~\cite{lalaloc++} & N/A & 2D Trans. & 0.72 & 0.92 \\
Laser2D~\cite{laser2d} & N/A & 2D Trans. + Rot. & 0.79 & 0.95 \\
Ours & Layout Lines & 3D Trans. + Rot. & \textbf{0.95} & \textbf{0.96} \\
\bottomrule
\end{tabular}
}
\vspace{-0.5em}
\caption{Localization evaluation using lines from indoor floorplans in Structured3D~\cite{Structured3D}. 
Note for baselines that only estimate translation, we report the accuracy using only the translation threshold.}
\vspace{-1.5em}
\label{table:layout_localization}
\end{table}

While our method is originally designed for localizing against 3D line maps obtained from point cloud scans or structure from motion, we find that our fully geometric setup is versatile to handle 3D floorplan maps without any hyperparameter changes.
We test our method on localizing panoramas against 3D lines extracted from a floorplan map~\cite{lalaloc,lalaloc++,laser2d} using the Structured3D~\cite{Structured3D} dataset.
Instead of simply extracting lines from panoramas, we use the lines from 2D layout annotations~\cite{Structured3D} to be compatible with the 3D floorplan maps.
Table~\ref{table:layout_localization} shows that our method can outperform existing methods.
Further, unlike prior works that only estimate the 2D translation and rotation, our method can estimate the full 6DoF pose while attaining competitive accuracy.
Therefore, if combined with sufficiently accurate 2D layout extraction from panoramas (removing the demand for 2D layout annotations), we expect our method to perform practical floorplan localization.

\section{Conclusion}
\vspace{-0.5em}
In this paper, we introduced an accurate and lightweight pipeline for fully geometric panoramic localization.
Our method is solely based on the geometry of lines, and thus can offer privacy protection while using a much smaller map size than methods using visual descriptors.
To effectively utilize the otherwise ambiguous geometric entities, we propose point distance functions along with an efficient comparison scheme for pose search and use principal directions of lines to match their intersections during refinement.
Due to the lightweight formulation, our method can perform scalable localization in large scenes and attain robustness in lighting changes.
We thus expect our method to serve as a practical pipeline for fully geometric localization.

\vspace{-1.5em}

\paragraph{Acknowledgements}
This work was supported by the National Research Foundation of Korea(NRF) grant funded by the Korea government(MSIT) (No. RS-2023-00218601) and and Samsung Electronics Co., Ltd. Young Min Kim is the corresponding author.

\appendix
\renewcommand\thetable{\thesection.\arabic{table}}    
\setcounter{table}{0}
\renewcommand\thefigure{\thesection.\arabic{figure}}    
\setcounter{figure}{0}
\setlength{\tabcolsep}{3pt}

\renewcommand{\theequation}{\thesection.\arabic{equation}}
\maketitlesupplementary
\section{Method Details}
\subsection{Input Preparation}
\label{sec:input_supp}
As explained in Section~\textcolor{red}{4.1}, our method operates in a fully geometric manner using lines and their intersections for localization.
Below we explain the detailed procedures for how the inputs are prepared prior to localization.

\paragraph{Line Extraction}
Similar to LDL~\cite{ldl}, our method extracts line segments in 2D and 3D using off-the-shelf line detectors.
For 2D, we create perspective crops of the input panorama and apply LSD~\cite{LSD}.
The detected lines are then converted to the spherical coordinate frame compatible to our method.
For 3D, we extract lines from the colored point cloud provided in OmniScenes~\cite{piccolo} and Stanford 2D-3D-S~\cite{stanford2d3d} by applying the 3D line detection algorithm from Xiaohu et al.~\cite{3d_lineseg}.

\paragraph{Principal Direction Extraction}
To imbue spatial context to lines, we additionally extract principal directions from lines as in LDL~\cite{ldl}.
For 2D, we find the principal directions by estimating vanishing points.
Specifically, we extrapolate all lines in 2D and find their intersections.
Then the intersection points are binned to a spherical grid, from which the top-3 bins with the largest number of points are selected and used as 2D principal directions.
For 3D, we similarly bin the line directions on the spherical grid and extract the top-3 directions.
Note we can apply more sophisticated vanishing point estimation algorithms such as Pautrat et al.~\cite{vanishing_point_pautrat}, which will lead to an enhanced localization performance since principal directions are first used for rotation estimation.
We leave such extensions to future work.

\paragraph{Line Intersection Extraction}
Our method uniquely leverages line intersections as important cues for accurate pose search and refinement.
As mentioned in Section~\textcolor{red}{4.1}, we intersect lines from distinct principal directions.
First for 2D line pairs, we extrapolate the lines on the sphere and obtain the intersection coordinates, and then keep intersections only if its spherical distance to both line segments are below a designated threshold $\delta_\text{2D}=0.1\text{rad}$.
Similarly for 3D line pairs, we extrapolate lines in the 3D space to get intersections, and then keep the intersections only if the distances to both line segments are below a threshold $\delta_\text{3D}=0.15m$.

\subsection{Theoretical Analysis of Efficient Distance Function Comparison}
In this section, we theoretically analyze the efficient distance function comparison presented in Section~\textcolor{red}{4.2.2}.
Recall that instead of exhaustively computing distance functions on-the-fly as in prior work~\cite{ldl}, we propose to cache distance function values prior to localization to enable scalable and efficient localization.
We start by proving Theorem 1 from Section~\textcolor{red}{4.2.2} which justifies our efficient distance function comparison.
The proof builds upon the following lemma:

\paragraph{Lemma 1.}
Given a metric $d(\cdot, \cdot)$ defined over the unit sphere $\mathbb{S}^2$, let $f(x; S){:=} \min_{s \in S} d(x, s)$ denote a distance function to a set of spherical points $S \! \subset \! \mathbb{S}^2$.
For two arbitrary points $p_1, p_2$ on the unit sphere $\mathbb{S}^2$, there exists a point $s^* \in S$ that satisfies the following inequality,
\begin{equation}
    |f(p_1; S) - f(p_2; S)| \leq |d(p_1, s^*) - d(p_2, s^*)|.
\label{eq:lemma}
\end{equation}

\paragraph{Proof}
Without loss of generality, suppose that $f(p_1; S) \geq f(p_2;S)$.
Also, let $s_i = \argmin_{s \in S} d(p_i, s)$ for $i=1,2$.
Then we have the following,
\begin{align}
    & |f(p_1; S) - f(p_2; S)| = d(p_1, s_1) - d(p_2, s_2) \\
    & \leq d(p_1, s_2) - d(p_2, s_2) \leq |d(p_1, s_2) - d(p_2, s_2)|,
\end{align}
and thus by setting $s^* = s_2$, Equation~\ref{eq:lemma} holds true, which proves the lemma.

\paragraph{}
Using the lemma, we prove the following theorem stated in Section~\textcolor{red}{4.2}:
\paragraph{Theorem 1.}
\textit{
Consider a countable, finite set of spherical points $Q \! \subset \! \mathbb{S}^2$ that satisfy $\max_{q \in Q}\min_{\hat{q} \in Q} d(q, R\hat{q}) \! \leq \! \max_{q \in Q}\min_{\hat{q} \neq q} d(q, \hat{q}) \! = \! \delta$ for all $R \in SO(3)$.
For an arbitrary rotation $\tilde{R} \in SO(3)$, the following inequality holds:
}
\begin{equation}
\frac{1}{|Q|}\sum_{q \in Q} |f(q; S) - f(\argmin_{\hat{q}\in Q} d(\hat{q}, \tilde{R}q); \tilde{R}S)|\leq \delta.
\label{eq:approx_bound}
\end{equation}

\paragraph{Proof}

We prove the following inequality for an arbitrary query point $q \in Q$, which when summed for all points in $Q$ equivalent to Equation~\ref{eq:approx_bound},
\begin{equation}
    |f(q; S) - f(\argmin_{\hat{q}\in Q} d(\hat{q}, \tilde{R}q); \tilde{R}S)|\leq \delta.
    \label{eq:single_term_bound}
\end{equation}
First, since distance function values are preserved under rotation of both the query point $q$ and the point set, we have $f(q; S) = f(\tilde{R}q; \tilde{R}S)$.
Further, Lemma 1 suggests that there exists a point $s^* \in S$ satisfying the following inequality,
\begin{align}
    & |f(\tilde{R}q; \tilde{R}S) - f(\argmin_{\hat{q}\in Q} d(\hat{q}, \tilde{R}q); \tilde{R}S)| \\
    & \leq |d(\tilde{R}q, \tilde{R}s^*) - d(\argmin_{\hat{q}\in Q} d(\hat{q}, \tilde{R}q), \tilde{R}s^*)|.
\end{align}
Using the triangle inequality and the dense point assumption $\max_{q \in Q}\min_{\hat{q} \in Q} d(q, \tilde{R} \hat{q}) \leq  \delta$, we have
\begin{align}
    & |d(\tilde{R}q, \tilde{R}s^*) - d(\argmin_{\hat{q}\in Q} d(\hat{q}, \tilde{R}q), \tilde{R}s^*)| \\
    & \leq d(\tilde{R}q, \argmin_{\hat{q}\in Q} d(\hat{q}, \tilde{R}q)) = \min_{\hat{q}\in Q}d(\hat{q}, \tilde{R}q) \leq \delta,
\end{align}
which in turn proves Equation~\ref{eq:single_term_bound}.

\paragraph{Derivation of Efficient Distance Function Comparison}
Using the theorem, we can derive the efficient distance function comparison presented in Section~\textcolor{red}{4.2.2}.
Given an arbitrary rotation $\tilde{R}$, from Equation~\ref{eq:approx_bound} we have that
\begin{align}
    &\frac{1}{|Q|} \sum_{q\in Q} |f(q;S) - f(\argmin_{\hat{q}\in Q} d(\hat{q}, \tilde{R}q); \tilde{R}S)| \\
    &= \frac{1}{|Q|} \sum_{q\in Q} |f(q;S) - f(\argmin_{\hat{q}\in \tilde{R}^TQ} d(\hat{q}, q); S)| \leq \delta.
\end{align}
This implies that for a one-to-one mapping $m(\cdot): Q \rightarrow \tilde{R}^TQ$ that satisfies $d(m(q),q)\leq \delta$ for all $q \in Q$, 
\begin{equation}
    \label{eq:one_to_one}
    \frac{1}{|Q|}\sum_{q\in Q} |f(q;S) - f(m(q); S)| \leq \delta.
\end{equation}
Note that the existence of such a mapping is given from the Hall's marriage theorem~\cite{hall_marriage}, assuming that for an arbitrary subset of points $\mathcal{Q} \subset Q$, we can find a subset of points $\mathcal{M} \subset \tilde{R}Q$ that satisfies $\max_{q \in \mathcal{Q}, m \in \mathcal{M}}d(q, m) \leq \delta$ and $|\mathcal{Q}| \leq |\mathcal{M}|$.

We can now use the results to derive a error bound on our approximation scheme. 
For line distance functions, the cumulative deviation between using our approximation (Equation~\textcolor{red}{8} from Section~\textcolor{red}{4.2.2}) and the original LDL~\cite{ldl} formulation (Equation~\textcolor{red}{3} from Section~\textcolor{red}{3}) is bounded as follows,
\begingroup
\allowdisplaybreaks
\begin{align}
    & \frac{1}{|Q|}\bigg |\underbrace{\sum_{q \in Q} |f_{2D}(q; L_{2D}) - f_{3D}(q; L_{3D}, R, t)|}_{\text{original (LDL)}} \\
    & - \underbrace{\sum_{q \in Q} |f_{2D}(q; R^TL_{2D}) - f_{3D}(q; L_{3D}, I, t)|}_{\text{ours}} \bigg | \\
    & =\frac{1}{|Q|}\bigg |\sum_{q \in Q} |f_{2D}(q; L_{2D}) - f_{3D}(q; L_{3D}, R, t)| \\
    & - \sum_{q \in Q} |f_{2D}(Rq; L_{2D}) - f_{3D}(Rq; L_{3D}, R, t)|\bigg | \\
    & \leq \frac{1}{|Q|}\bigg | \sum_{q \in Q} f_{2D}(q; L_{2D}) - f_{3D}(q; L_{3D}, R, t) \\
    & - \sum_{q \in Q} f_{2D}(Rq; L_{2D}) + f_{3D}(Rq; L_{3D}, R, t)\bigg | \\
    & = \frac{1}{|Q|}\bigg | \sum_{q \in Q} \big (f_{2D}(q; L_{2D}) - f_{2D}(m(q); L_{2D}) \big)\\
    & + \big(f_{3D}(m(q); L_{3D}, R, t) - f_{3D}(q; L_{3D}, R, t)\big)\bigg | \\
    & \leq \frac{1}{|Q|} \sum_{q \in Q} \big (|f_{2D}(q; L_{2D}) - f_{2D}(m(q); L_{2D})|\\
    & + |f_{3D}(m(q); L_{3D}, R, t) - f_{3D}(q; L_{3D}, R, t)|\big ) \leq 2\delta,
\end{align}
\endgroup
where $m(\cdot): Q \rightarrow RQ$ is the one-to-one mapping defined similarly as in Equation~\ref{eq:one_to_one}.
Therefore, when we have a sufficiently small $\delta$ from a dense set of query points $Q$, our efficient approximation closely follows the original line distance function comparison from LDL.
A similar derivation can also be made for point distance functions, but note that for point distance functions we have the sharpening parameter $\gamma=0.2$ applied to the spherical distance (Equation~\textcolor{red}{4}, \textcolor{red}{5}).
Theorem 1 and its consequences still hold however, due to the fact that for a metric $d(\cdot, \cdot)$ defined on a bounded set, the composition of the metric with an increasing concave function $f(\cdot)$, namely $f(d(\cdot, \cdot))$ is still a metric~\cite{snowflake_1,snowflake_2} (also known as the \textit{snowflake} metric). 

\subsection{Translation / Rotation Pool Generation}
We further elaborate on how the translation and rotation pools are generated for our method.
As explained in Section~\textcolor{red}{3}, we follow the pool generation method from LDL~\cite{ldl}.
Specifically, we generate translation pools by first creating a bounding box of the 3D map and subdividing the bounding box into grid partitions.
The center of each grid is used as the translation pool.
We then generate rotation pools by combinatorially associating principal directions in 2D and 3D.
Given three principal directions in 2D and 3D, namely $d_i \in D_{2D}$ and $\tilde{d}_i \in D_{3D}$, each rotation is determined from a permutation that associates directions in 2D with 3D.
To elaborate, a rotation matrix can be obtained from an arbitrary permutation $\sigma(\cdot)$ by applying the Kabsch algorithm~\cite{kabsch} on direction pairs $\{(d_1, \tilde{d}_{\sigma(1)}), (d_2, \tilde{d}_{\sigma(2)}), (d_3, \tilde{d}_{\sigma(3)})\}$.
Note that due to the additional sign ambiguity when associating the principal directions (i.e., for a fixed $\sigma(\cdot)$, $d_i$ can be associated with $\pm d_{\sigma(i)}$), there can exist $2^3 \times 3\!$ associations.

\subsection{Hyperparameter Setup}
We report details about the hyperparameters not reported in the main paper.
First for pose search, we filter short lines to reduce the effect of noisy line misdetections.
We specifically filter 3D lines whose lengths are over $20cm$, and then filter 2D lines by length to match the ratio of filtered 3D lines.
For pose refinement, we use Adam~\cite{adam} for optimizing the cost functions (Equation~\textcolor{red}{14}, ~\textcolor{red}{15}), with a step size of $0.1$ for $100$ iterations.

\section{Additional Experimental Results}
\begin{figure}[t]
  \centering
    \includegraphics[width=\linewidth]{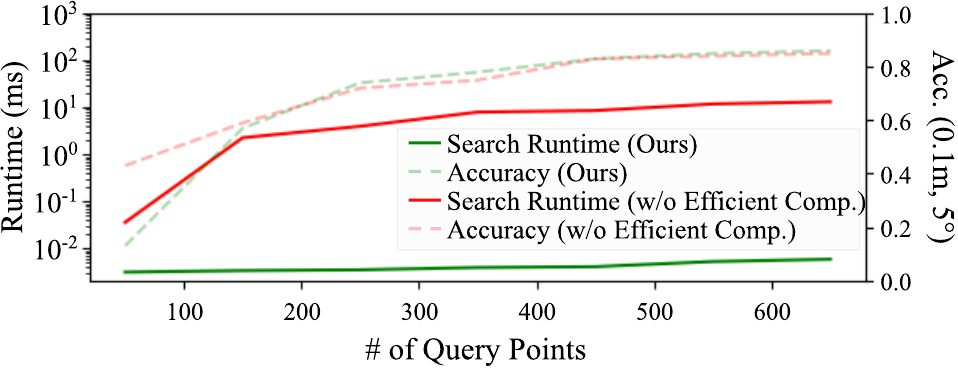}
   \caption{Efficacy of our efficient distance function comparison on localization accuracy and pose search runtime. We plot the metrics under varying number of query points $|Q|$ on Room 5 from OmniScenes~\cite{piccolo}. Larger number of query points lead to significantly enhanced accuracy, but the distance function comparison originally propose from LDL~\cite{ldl} shows large increase in runtime. On the other hand, our formulation shows an almost constant pose search runtime with nearly no loss in localization accuracy. Note the runtime is plotted in log scale.}
   \label{fig:ablation_query_points}
\vspace{-1em}
\end{figure}
\subsection{Scalability against Number of Query Points}
One of the key factors for accurate localization of our method is the dense set of query points used during pose search.
In all our experiments, we use $|Q|=642$ query points, which is much larger than that of LDL ($|Q|=42$).
By using denser query points, we can better compare the fine-grained scene details, leading to enhanced pose search performance.
However, naively increasing the number of query points leads to a significant increase in pose search runtime.

The efficient distance function comparison enables our method to handle dense query points without a noticeable increase in runtime.
To illustrate the effect of using our efficient formulation, we assess the runtime and localization accuracy in Room 5 from OmniScenes~\cite{piccolo} using varying number of query points.
Figure~\ref{fig:ablation_query_points} plots the accuracy and pose search runtime with respect to the number of query points.
First, one can observe that there is a clear, positive correlation between the number of query points and localization accuracy.
However, without the efficient comparison scheme, runtime also dramatically increases.
As our distance function comparison scheme pre-computes and caches 3D distance functions, which is the largest bottleneck for scaling query points, pose search runtime remains almost constant with the increase in query points.
Further, note that one can attain a reasonable localization accuracy once query points are sufficiently dense ($|Q| \geq 350$ for Figure~\ref{fig:ablation_query_points}).
Therefore, for memory critical applications, one may choose to employ a smaller number of query points to reduce the map size.

\begin{table}[]
\centering
\resizebox{0.8\linewidth}{!}{
\begin{tabular}{lccc}
\toprule
Method & \textit{t}-error & \textit{R}-error & \begin{tabular}[c]{@{}c@{}}Accuracy\\ (0.1m, $5^\circ$)\end{tabular} \\
\midrule
Only rot. refine & 0.21 & 1.56 & 0.08 \\
Alternating trans. \& rot. refine & 0.07 & 1.55 & 0.74 \\
No intersection match update & 0.17 & 1.56 & 0.13 \\
Ours & \textbf{0.06} & \textbf{1.05} & \textbf{0.77} \\
\bottomrule
\end{tabular}
}
\caption{Ablation study of key components of our pose refinement method, using the Room 2 subset from OmniScenes~\cite{piccolo}.}
\label{table:abl_supp}
\vspace{-1em}
\end{table}

\subsection{Additional Ablation Study for Pose Refinement}
In this section we conduct additional ablation experiments for the pose refinement module.
Recall from Section~\textcolor{red}{4.3} that we refine poses by aligning line intersections on the sphere, where translation is first optimized followed by rotation.
We consider three ablated variations of pose refinement: (i) only optimizing rotation, (ii) alternating translation and rotation refinement each step, and (iii) omitting match updates during each translation refinement step.
The results are shown in Table~\ref{table:abl_supp}, where we conduct evaluation in the identical setup as in Section~\textcolor{red}{5.3}.
Only optimizing rotation leads to a large drop in localization accuracy, since the translations from initial pose pools are usually at least $0.10m$ away from the ground truth.
Further, alternating translation and rotation instead of optimizing them sequentially lead to a slight drop in performance.
As the initial rotation estimate is already fairly accurate (obtained by aligning principal directions), placing more weight on optimizing translation during the initial stage of refinement is beneficial.
Finally, omitting the match update scheme leads to a dramatic decrease in performance.
As the initial translations are imperfect, the iterative updates during translation refinement perform a crucial role in both obtaining good translations and intersection point matches.

\subsection{Robustness Against Line Detector Variations}
We further evaluate the robustness of our method against line detector variations.
As mentioned in Section~\ref{sec:input_supp}, we apply LSD~\cite{LSD} for 2D line detection.
In this section we test if our method can also handle line segments from other detection methods.
We conduct evaluations in the OmniScenes dataset~\cite{piccolo}, using the top-1 retrieval results for refinement.
To fairly test the generalizability of our method, all the hyperparameters are fixed to the setup used for the original set of detectors.

We test variations in 2D line detectors using two recently proposed methods: ELSED~\cite{elsed} and DeepLSD~\cite{deeplsd}.
Similar to how we applied LSD~\cite{LSD} on panoramas, we first make perspective crops of the panorama and apply the detectors.
As shown in Table~\ref{table:line_detector_var}, our method shows consistent performance amidst changes in the 2D line detectors.
The results suggest that our formulation is sufficiently generalizable and versatile to handle varying line detectors.

\begin{table}[]
\centering
\resizebox{0.65\linewidth}{!}{
\begin{tabular}{lccc}
\toprule
Line Detector & \textit{t}-error & \textit{R}-error & \begin{tabular}[c]{@{}c@{}}Accuracy\\ (0.1m, $5^\circ$)\end{tabular} \\
\midrule
ELSED~\cite{elsed} & \textbf{0.06} & 0.80 & 0.71 \\
DeepLSD~\cite{deeplsd} & \textbf{0.06} & \textbf{0.73} & 0.71 \\
LSD~\cite{LSD} (Ours) & \textbf{0.06} & 0.96 & \textbf{0.77} \\
\bottomrule
\end{tabular}
}
\caption{Robustness evaluation against line detector variations, using the Extreme split from OmniScenes~\cite{piccolo}.}
\label{table:line_detector_var}
\vspace{-1em}
\end{table}

\begin{table*}[t]
\centering
\resizebox{0.85\linewidth}{!}{


\begin{tabular}{llc|ccc|ccc}
\toprule
\multicolumn{3}{l}{Dataset} & \multicolumn{3}{c}{OmniScenes} & \multicolumn{3}{c}{Stanford 2D-3D-S} \\
\midrule
Refinement & Method & \multicolumn{1}{l|}{\begin{tabular}[c]{@{}l@{}}Visual\\ Desc.\end{tabular}} & \begin{tabular}[c]{@{}c@{}}Accuracy \\ (0.1m, $5^\circ$)\end{tabular} & \begin{tabular}[c]{@{}c@{}}Accuracy \\ (0.2m, $10^\circ$)\end{tabular} & \begin{tabular}[c]{@{}c@{}}Accuracy \\ (0.3m, $15^\circ$)\end{tabular} & \begin{tabular}[c]{@{}c@{}}Accuracy \\ (0.1m, $5^\circ$)\end{tabular} & \begin{tabular}[c]{@{}c@{}}Accuracy \\ (0.2m, $10^\circ$)\end{tabular} & \begin{tabular}[c]{@{}c@{}}Accuracy \\ (0.3m, $15^\circ$)\end{tabular} \\

\midrule
\multirow{2}{*}{\begin{tabular}[c]{@{}l@{}}Line-Based \\ Refinement\end{tabular}} & Line Transformer~\cite{line_transformer} & $\bigcirc$ & 0.88 & 0.92 & 0.93 & 0.70 & \textbf{0.72} & \textbf{0.73} \\
 & Gluestick~\cite{gluestick} & $\bigcirc$ & 0.89 & 0.93 & 0.94 & 0.68 & 0.71 & 0.72 \\

\midrule
\multirow{5}{*}{\begin{tabular}[c]{@{}l@{}}Point-Based \\ Refinement\end{tabular}} & SuperGlue~\cite{sarlin2020superglue} & $\bigcirc$ & 0.90 & \textbf{0.95} & \textbf{0.95} & \textbf{0.71} & \textbf{0.72} & \textbf{0.73} \\
 & LightGlue~\cite{lightglue} & $\bigcirc$ & \textbf{0.93} & \textbf{0.95} & \textbf{0.95} & \textbf{0.71} & \textbf{0.72} & \textbf{0.73} \\
 & LoFTR~\cite{loftr} & $\bigcirc$ & 0.88 & 0.94 & \textbf{0.95} & 0.67 & 0.69 & 0.69 \\
 & PICCOLO~\cite{piccolo} & $\bigcirc$ & 0.58 & 0.60 & 0.61 & 0.57 & 0.59 & 0.60 \\
 & CPO~\cite{cpo} & $\bigcirc$ & 0.76 & 0.78 & 0.78 & 0.55 & 0.56 & 0.58 \\

\midrule
\multirow{10}{*}{\begin{tabular}[c]{@{}l@{}}Geometric \\ Refinement\end{tabular}} & GoMatch~\cite{gomatch} & \xmark & 0.67 & 0.84 & 0.88 & 0.57 & 0.65 & 0.67 \\
 & BPnPNet~\cite{bpnp_net} & \xmark & 0.01 & 0.18 & 0.41 & 0.05 & 0.24 & 0.41 \\
 & Gao et. al~\cite{line_refinement} & \xmark & 0.11 & 0.43 & 0.74 & 0.26 & 0.63 & 0.62 \\
 & PDF Minimization & \xmark & 0.28 & 0.52 & 0.63 & 0.18 & 0.31 & 0.38 \\
 & LoFTR\textsuperscript{L}~\cite{loftr} & \xmark & 0.06 & 0.26 & 0.55 & 0.10 & 0.33 & 0.47 \\
 & SuperGlue\textsuperscript{L}~\cite{sarlin2020superglue} & \xmark & 0.06 & 0.28 & 0.60 & 0.06 & 0.37 & 0.56 \\
 & LightGlue\textsuperscript{L}~\cite{lightglue} & \xmark & 0.45 & 0.69 & 0.82 & 0.33 & 0.51 & 0.58 \\
 & GlueStick\textsuperscript{L}~\cite{gluestick} & \xmark & 0.36 & 0.60 & 0.75 & 0.30 & 0.46 & 0.55 \\
 & Line Transformer\textsuperscript{L}~\cite{line_transformer} & \xmark & 0.05 & 0.24 & 0.54 & 0.07 & 0.31 & 0.47 \\
 & Ours & \xmark & 0.77 & 0.89 & 0.91 & 0.62 & 0.66 & 0.67 \\
 \bottomrule
\end{tabular}

}

\caption{Pose refinement evaluation in OmniScenes~\cite{piccolo} and Stanford 2D-3D-S~\cite{stanford2d3d}. We retrieve top-1 poses using point distance functions and perform refinement with various baseline methods. Note the superscript X\textsuperscript{L} denotes that the baseline takes line images as input.}
\label{table:pose_refine_full}

\end{table*}

\begin{table*}[t]
\centering
\resizebox{0.8\linewidth}{!}{


\begin{tabular}{lc|ccccccc|cc}
\toprule
Method & \multicolumn{1}{l|}{\begin{tabular}[c]{@{}l@{}}Visual\\ Desc.\end{tabular}} & Orig. & Intensity\textsuperscript{1} & Gamma\textsuperscript{1} & \begin{tabular}[c]{@{}c@{}}White \\ Balance\textsuperscript{1}\end{tabular} & Intensity\textsuperscript{2} & Gamma\textsuperscript{2} & \begin{tabular}[c]{@{}c@{}}White \\ Balance\textsuperscript{2}\end{tabular} & Range & Std \\

\midrule
Line Transformer~\cite{line_transformer} & $\bigcirc$ & 0.88 & 0.67 & 0.88 & \textbf{0.89} & 0.73 & 0.83 & 0.89 & 0.22 & 0.09 \\
Gluestick~\cite{gluestick} & $\bigcirc$ & \textbf{0.89} & \textbf{0.77} & \textbf{0.89} & \textbf{0.89} & \textbf{0.83} & \textbf{0.89} & \textbf{0.90} & 0.13 & 0.05 \\
\midrule
Line Transformer~\cite{line_transformer} & \xmark & 0.67 & 0.64 & 0.66 & 0.67 & 0.68 & 0.60 & 0.66 & 0.08 & 0.03 \\
Gluestick~\cite{gluestick} & \xmark & 0.36 & 0.47 & 0.45 & 0.30 & 0.51 & 0.29 & 0.31 & 0.22 & 0.09 \\
Ours & \xmark & 0.77 & 0.72 & 0.75 & 0.77 & 0.75 & 0.72 & 0.77 & \textbf{0.05} & \textbf{0.02}\\
\bottomrule
\end{tabular}

}

\caption{Pose refinement evaluation under varying lighting conditions in OmniScenes~\cite{piccolo}. We report the localization accuracy at 0.1m and $5^\circ$, along with their range and standard deviations. Note we test two levels of variations for each lighting change type, totalling six variations.}
\label{table:lighting_refine_full}

\end{table*}

\subsection{Full Experimental Results}
We finally report the full evaluation results for pose refinement in Stanford 2D-3D-S~\cite{stanford2d3d} and OmniScenes~\cite{piccolo}.
Note that the results presented in Table~\textcolor{red}{3} and~\textcolor{red}{5} are the results aggregated from the two datasets.
Table~\ref{table:pose_refine_full} and~\ref{table:lighting_refine_full} shows the full results for pose refinement in regular setups and lighting variations.
In both cases, our method shows competitive performance against the visual descriptor-based methods while constantly outperforming the geometry-based methods.

\section{Baseline Details}
In this section, we explain the implementation details of the baselines compared against our method.

\paragraph{Pose Search Baselines}
We consider three types of baselines for comparison: neural network based (SFRS~\cite{openibl}, Cosplace~\cite{cosplace}), color distribution-based (PICCOLO~\cite{piccolo}, CPO~\cite{cpo}), and line-based (LDL~\cite{ldl}, Chamfer~\cite{line_chamfer}, SFRS\textsuperscript{L}, Cosplace\textsuperscript{L}).
Neural network-based methods operate by first extracting global feature descriptors for image renderings in the 3D map and establishing comparisons against that obtaine from the query image.
We specifically use the colored point clouds available in our test datasets to render synthetic views.
Color distribution-based methods operate by directly comparing the color values between the panorama and point cloud.
PICCOLO~\cite{piccolo} operates using a loss function defined over various candidate poses that measures the color discrepancy between the point cloud projections and the panorama color values sampled from the projection locations.
CPO~\cite{cpo} takes a more holistic approach by comparing the patch-level color histograms of the query and synthetic views in the point cloud.
Line-based methods solely utilize the geometry of the line maps for pose search.
LDL~\cite{ldl} uses line distance functions for pose search as explained in Section~\textcolor{red}{3}, while Chamfer-based method~\cite{line_chamfer} uses the pairwise distances between the lines in 2D and 3D for pose search.
We additionally test variants of the neural network-based methods, namely SFRS\textsuperscript{L} and Cosplace\textsuperscript{L}, that extracts global descriptors from the 2D line extractions and those from synthetic views in 3D.
As demonstrated in Section~\textcolor{red}{5.2}, our method performs competitively against the baselines by leveraging the holistic geometry of lines and their intersections, while exhibiting a very short runtime due to the efficient search scheme.

\paragraph{Pose Refinement Baselines}
Similar to pose search evaluation, we consider three types of baselines: line-based (GlueStick~\cite{gluestick}, Line Transformer~\cite{line_transformer}), point-based (SuperGlue~\cite{sarlin2020superglue}, LightGlue~\cite{lightglue}, LoFTR~\cite{loftr}, PICCOLO~\cite{piccolo}, CPO~\cite{cpo}), and geometry-based (GoMatch~\cite{gomatch}, BPnPNet~\cite{bpnp_net}, Gao et. al~\cite{line_refinement}, PDF Minimization, LoFTR\textsuperscript{L}, SuperGlue\textsuperscript{L}, LightGlue\textsuperscript{L}, GlueStick\textsuperscript{L}, Line Transformer\textsuperscript{L}).
Line-based methods operate by matching visual descriptors for lines and additionally for points, where we apply PnL-RANSAC~\cite{cvxpnpl,ransac} similar to Yoon et al.~\cite{line_transformer} on the point and line matches to obtained refinement results.

Point-based methods can be further divided into neural network-based methods (SuperGlue~\cite{sarlin2020superglue}, LightGlue~\cite{lightglue}, LoFTR~\cite{loftr}) and color matching-based methods (PICCOLO~\cite{piccolo}, CPO~\cite{cpo}).
Neural network-based methods match point descriptors using graph neural networks or transformers, where we apply PnP-RANSAC~\cite{pnp_1, pnp_2, ransac} on the matches to get a refined pose.
Note for LightGlue~\cite{lightglue} we used the SuperPoint~\cite{superpoint} keypoint descriptors as input to the matcher.
Both color matching-based methods tested in our experiments minimize a loss function termed sampling loss~\cite{piccolo}, which directly compares the point color values against the panorama image's color values at projected locations.
Here, the distinction between PICCOLO and CPO is in that CPO additionally exploits 3D score maps to place weights on regions less likely to contain changes during sampling loss optimization.

Finally, for geometry-based methods we specifically test neural network-based methods (GoMatch~\cite{gomatch}, BPnPNet~\cite{bpnp_net}),   optimization-based methods (Gao et al.~\cite{line_refinement}, PDF minimization), and line image-based methods (LoFTR\textsuperscript{L}, SuperGlue\textsuperscript{L}, LightGlue\textsuperscript{L}, GlueStick\textsuperscript{L}, Line Transformer\textsuperscript{L}).
For neural network-based methods which operate by matching learned descriptors for keypoint locations, we use SuperPoint~\cite{superpoint} keypoint detections.
In addition, as the bearing vector representation of GoMatch cannot fully handle the $360^\circ$ view of panoramas, we subdivide the panoramas into $N_\text{split}$ horizontally split regions and separately apply GoMatch.
We set $N_\text{split}=8$ in all our experiments, as this attained the highest performance.
For optimization-based methods, first Gao et al.~\cite{line_refinement} matches lines geometrically by inspecting the amount of line overlaps and then refines the matches by aligning line midpoint distances and directions.
PDF minimization is a conceived baseline to test if point distance functions can be further used for refining poses.
Here we apply gradient descent optimization on the cost function from Equation~\textcolor{red}{6}, using Adam~\cite{adam} with a step size of 0.1.
Similar to pose search evaluation, the line image-based methods considered here also take the line images in 2D and 3D, and apply visual descriptor-based matching to obtain the refined pose.

\section{Details on Experiment Setup}

\paragraph{Lighting Robustness Evaluation}
We elaborate on the details of the lighting changes in Section~\textcolor{red}{5.2}, which are used to evaluate the illumination robustness of our method.
As shown in Table~\ref{table:lighting_refine_full}, we apply three types of color variations to images in the entire Extreme split of OmniScenes~\cite{piccolo}: average intensity, gamma, and white balance change.
For average intensity, we lower the pixel intensity by $25\%$ (Intensity\textsuperscript{1}) and $33\%$ (Intensity\textsuperscript{2}).
For gamma, we test image gamma values with 0.3 (Gamma\textsuperscript{1}) and 1.5 (Gamma\textsuperscript{2}).
For white balance, we apply the following linear transformations to the RGB values: $\begin{pmatrix}
0.9 & 0 & 0\\
0 & 0.5 & 0\\
0 & 0 & 0.7
\end{pmatrix}$ (White Balance\textsuperscript{1}), $\begin{pmatrix}
0.6 & 0 & 0\\
0 & 0.9 & 0\\
0 & 0 & 0.4
\end{pmatrix}$ (White Balance\textsuperscript{2}).

\begin{figure}[t]
  \centering
    \includegraphics[width=\linewidth]{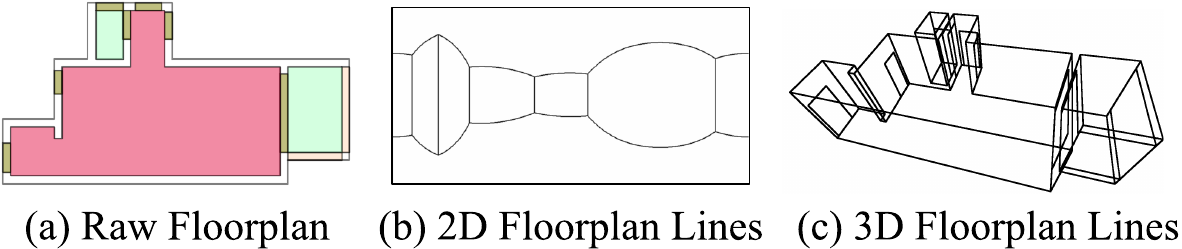}
   \caption{
   Visualization of lines in 2D and 3D extracted from the floorplans in Structured3D~\cite{Structured3D} dataset. Given the raw floorplan annotations and the room height information, we extract the 2D and 3D floorplan lines.
   }
   \label{fig:floorplan}
\vspace{-1em}
\end{figure}

\paragraph{Floorplan Localization Evaluation}
To test the generalizability of our method against sparser line maps, we evaluated localization performance using floorplans in Section~\textcolor{red}{5.4}.
Here we describe the details on the experiment setup.
As shown in Figure~\ref{fig:floorplan}, from the raw floorplan annotations and known room height, we extract the 2D and 3D lines.
Then, without any modifications in the hyperparameter setup, the lines are given as input to our localization pipeline.
Despite any floorplan-specific tuning, our method performs competitively against the state-of-the-art methods, as demonstrated in Section~\textcolor{red}{5.3}.

{
    \small
    \bibliographystyle{ieeenat_fullname}
    \bibliography{main}
}

\end{document}